\documentclass{article}
\usepackage{spconf,amsmath,graphicx}
\usepackage{color}

\DeclareMathOperator*{\argmin}{arg\,min}
\def\by{\mathbf{y}}
\def\bx{\mathbf{x}}
\def\bw{\mathbf{w}}
\def\bh{\mathbf{h}}
\def\bv{\mathbf{v}}
\def\DD{\mathcal{D}} 
\def\LL{\mathcal{L}}

\newsavebox\CBox
\def\textBF#1{\sbox\CBox{#1}\resizebox{\wd\CBox}{\ht\CBox}{\textbf{#1}}}

\title{Personalized Federated Learning on  Long-Tailed   Data\\  via Adversarial Feature Augmentation}

\name{Yang Lu$^1$, Pinxin Qian$^1$, Gang Huang$^2$, Hanzi Wang$^{1,*}$\thanks{* Corresponding Author: Hanzi Wang, hanzi.wang@xmu.edu.cn}}
\address{$^1$Fujian Key Laboratory of Sensing and Computing for Smart City, 
Xiamen University, Xiamen, China\\ $^2$Zhejiang Lab, Hangzhou, Zhejiang, China \\pxqian@stu.xmu.edu.cn, huanggang@zju.edu.cn,  \{luyang,~hanzi.wang\}@xmu.edu.cn}

\begin{document}
%
\maketitle
\begin{abstract}
	Personalized Federated Learning (PFL) aims to learn personalized models for each client based on the knowledge across all clients in a privacy-preserving manner. Existing PFL methods generally assume that the underlying global data across all clients are uniformly distributed without considering the long-tail distribution. The joint problem of data heterogeneity and long-tail distribution in the FL environment is more challenging and severely affects the performance of personalized models. In this paper, we propose a PFL method called Federated Learning with Adversarial Feature Augmentation (FedAFA) to address this joint problem in PFL. FedAFA optimizes the personalized model for each client by producing a balanced feature set to enhance the local minority classes. The local minority class features are generated by transferring the knowledge from the local majority class features extracted by the global model in an adversarial example learning manner. The experimental results on benchmarks under different settings of data heterogeneity and long-tail distribution demonstrate that FedAFA significantly improves the personalized performance of each client compared with the state-of-the-art PFL algorithm. The code is  available at \color{blue}https://github.com/pxqian/FedAFA\color{black}.
\end{abstract}
\vspace{-10px}
\section{Introduction}
\label{sec:intro}

A common way to build a deep learning model is to collect the training data and train the model on a server, called centralized training. 
However, with the increasing data security awareness, such a centralized training paradigm is not applicable in some applications when sensitive data is stored in each data holder. Federated Learning (FL)  develops a paradigm to train models without transmitting private data from each data holder (called client in FL) to a centralized server to address the privacy issue in deep learning \cite{guliani2021training, cui2021federated,abad2020hierarchical,sami2022over}. However, one major problem in FL is that the aggregated global model is usually not guaranteed to generalize overall clients well because each client's data distribution is not-independent and identically distributed (non-IID), which is usually defined as the \textit{data heterogeneity problem} in FL. Therefore, targeting the model generalization ability on each client, Personalized Federated Learning (PFL) aims to obtain a `tailored' local model that utilizes the global model's generalization ability and simultaneously fits the client's local data distribution.
  
Existing PFL methods  \cite{l2sgd, pfedme,li2021ditto,maml} have generally achieved promising personalized performance of each client on heterogeneous data. However, the data usually exhibits long-tail distribution in real-world scenarios, where a few head classes contain most samples while a large number of tail classes contain only a few samples. Therefore, it is reasonable to assume that the global data (data across all clients) is in long-tail distribution. In the PFL environment with data heterogeneity, the global tail class samples will be sporadically distributed on only a few clients, and each client's data distribution is also locally imbalanced. In addition, the local data distributions are likely to differ from the global data distribution, which yields different data imbalance degrees among clients. In this case, the generalization ability of PFL models will further deteriorate because the global tail classes are under-represented by the aggregated global model, which mainly provides the knowledge of the global head classes. In addition, the PFL models are prone to overfitting the local minority classes with only a few samples in personalized optimization on each client. Some existing long-tail learning methods \cite{ cui2019class, Park_2021_ICCV, cao2019learning} are also not applicable in the PFL environment because the global data distribution is unknown to both the server and clients due to privacy issues.

To address the joint problem of data heterogeneity and long-tail distribution in PFL, we propose a PFL method called Federated Learning with Adversarial Feature Augmentation (FedAFA) that utilizes the global model learned across clients to rebalance the local feature set for robust personalized training. Specifically, inspired by the targeted adversarial attack \cite{ilyas2019adversarial,goodfellow2014explaining}, we generate new local minority class features by adding specific small perturbations to the local majority
class features. To not affect the performance of the original local majority classes, we also propose a new optimization objective during personalized training. Experimental results on various heterogeneous and long-tailed benchmarks show that FedAFA surpasses the state-of-the-art PFL and long-tail learning methods.

\vspace{-10px}

\section{Proposed Method}
\label{sec:pagestyle}
\vspace{-5px}

\subsection{Basics of Personalized Federated Learning}
In a typical PFL setting, there are $K$ clients participating in the training process and each client $k$ has a private dataset $\DD_k=\{(\bx_i,\by_i)\}_{i=1}^{n_k}$, where $\bx_i$ denotes the $i$th sample in $\DD_k$, $\by_i \in \{0,1\}^C$ is the corresponding label over $C$ classes, and $n_k$ is the number of samples in $\DD_k$. PFL aims to look for good local models (personalized models) for all $K$ clients, which are usually adapted from the global model $\bw$.  FedAvg-FT \cite{wang2019federated} is the most straightforward PFL method.  It is a locally adaptive algorithm based on fine-tuning the global model by $\bw_k=\bw-\eta \nabla \LL_k(\DD_k;\bw)$ on $\DD_k$, where $\LL_k(\DD_k;\bw)$ is the local training loss of the model $\bw$ on $\DD_k$. $\LL_k(\DD_k;\bw)=\frac{1}{|\DD_k|}\sum_{(\bx,\by)\in\DD_k}\ell(\by,f(\bx;\bw))$ is calculated by averaging sample losses in $\DD_k$,
where $\ell(\cdot,\cdot)$ is the sample loss and $f(\bx;\bw)$ is the prediction result of sample $\bx$ by model $\bw$. However, this method slightly improves heterogeneous data because local model $\bw_k$ is prone to overfit the local data $\DD_k$, which results in poor local generalization performance.
	\vspace{-10px}

\subsection{FedAFA Framework}\label{fedafa}
FedAFA is based on the following intuitions. First, the global model of FL is usually more robust than local models because it obtains the information from each client, although it is in an indirect manner. Second, it has been empirically shown that the feature extractor in a neural network is less affected by the data distribution compared with its classifier \cite{zhou2020bbn, Kang2020Decoupling,luo2021no}. It means that the features extracted by the global model are still of high quality despite the negative influence of data heterogeneity and long-tail distribution. Based on the above observations, we propose to utilize the feature extractor of the global model to transfer information from the local majority classes to the local minority classes in the feature space.

\begin{figure}[!t]
        \centering
        \includegraphics[width=0.45\textwidth]{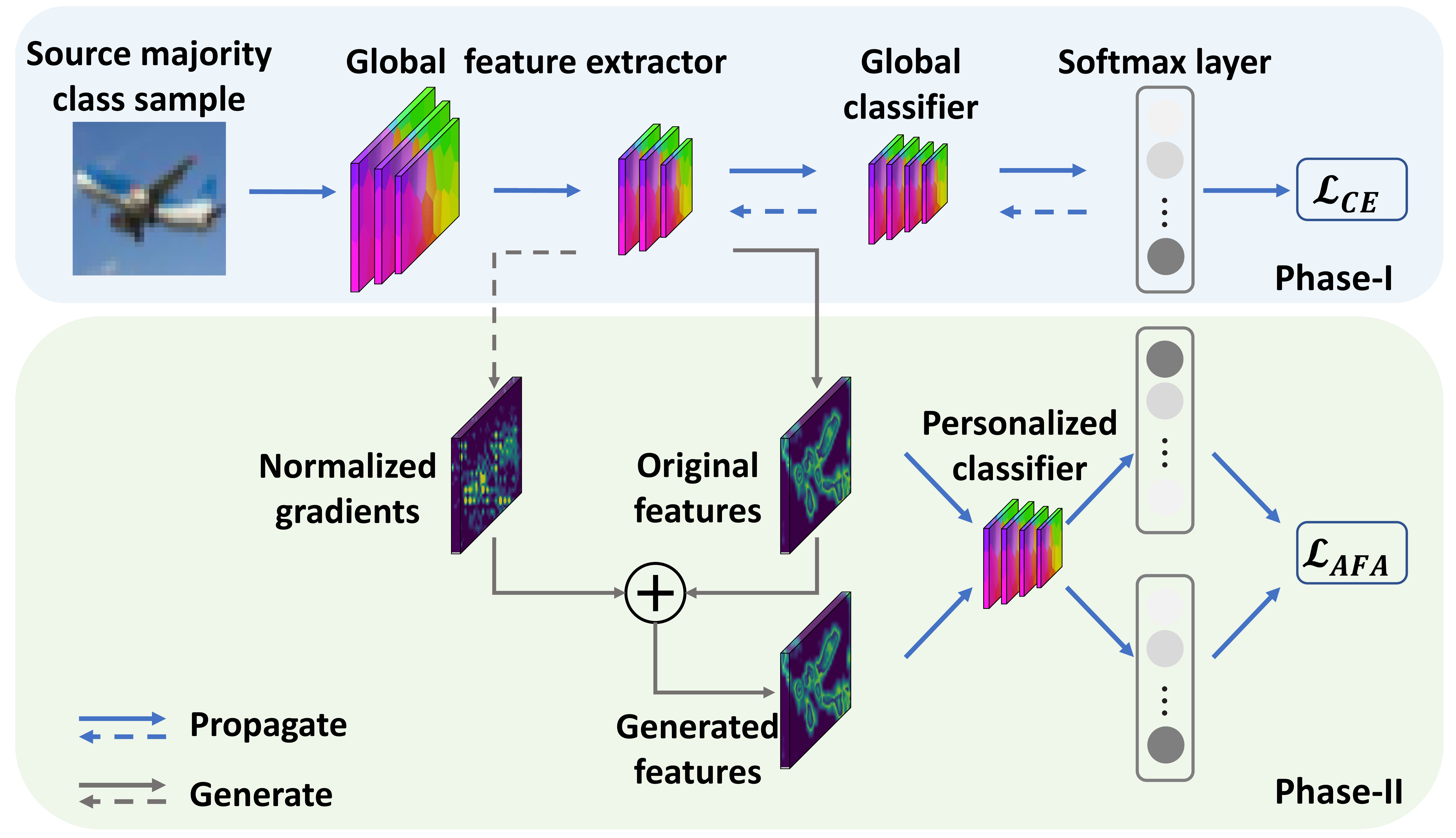} 
        \caption{An overview of personalized training of proposed FedAFA.} \label{overview}
		\vspace{-10px}
    \end{figure}

    We randomly select pairs of classes as the \textit{source majority class} $y_s$ and the \textit{target minority class} $y_t$ depending on their corresponding number of samples denoted as $n_s$ and $n_t$. We use the idea of adversarial samples \cite{ilyas2019adversarial,goodfellow2014explaining} to generate the features of the target minority class by adding specific perturbations to the features of source majority class samples. The perturbations are obtained from the gradients of the loss of predicting the source majority class sample into the target minority class. However, unlike generating adversarial samples, the input and output of FedAFA are in the feature space. After generating a certain number of features to achieve a balanced feature set for all local classes, we combine them with the original local samples with class-balanced sampling to train the local personalized model. Thus, the PFL model of FedAFA is constructed by the feature extractor of the global model and the personalized classifier trained on the rebalanced feature set, which is illustrated in Figure \ref{overview}. 
	In order to explain FedAFA with better notations, we split the model into two parts: a feature extractor $g$ parameterized by $\mathbf{u}$ and a classifier $f$ parameterized by $\mathbf{v}$. Thus, the features of a sample $\bx$ is generated by $\mathbf{h}=g(\bx;\mathbf{u})$, and the prediction result is given by $f(\mathbf{h};\mathbf{v})$. In FedAFA,  client $k$ uses the feature extractor of the global model $\mathbf{u}$ and has its own personalized classifier with parameters $\mathbf{\bv}_k$. Next, two major steps in FedAFA, i.e., local feature augmentation and personalized model optimization, are described in detail.
\vspace{-10px}
\subsubsection{Local Feature Augmentation}\label{fea} 
First, we introduce how to adversarially generate features of the target minority class by utilizing the information of the source majority class. Given a target minority class $\by_t$, the probability $p(\by_s|\by_t)$ of selecting the source majority class $\by_s$ is proportional to $Bernoulli (\frac{n_s-n_t}{n_k})$. Once a source majority class $\by_s$ is determined, a sample $(\bx_s,\by_s)\in\DD_k$ is randomly selected. Then, the sample is fed into the feature extractor of the global model to obtain its original features $\mathbf{h}_s=g(\bx_s;\mathbf{u})$. The original features $\mathbf{h}_s$  is going to be transformed into generated features $\widehat{\bh}_t$ in the target class $\by_t$. The objective of feature transformation is formulated as:
\vspace{-5px}
\begin{align}
 \widehat{\bh}_t=\argmin_{\widehat{\bh}_t:=\mathbf{h}_s+\delta} \ell(\by_t,f(\widehat{\bh}_t;\mathbf{\bv})),\label{e6}
\end{align}
where $\delta$ denotes the perturbation added to the original features $\mathbf{h}_s$, such that $\mathbf{h}_s+\delta$ can be predicted as the target minority class $\by_t$ by minimizing the loss $\ell(\by_t,f(\widehat{\bh}_t;\mathbf{\bv}))$. An intuitive strategy to optimize Eq. (\ref{e6}) is to obtain the perturbations $\delta$ by the negative normalized gradients of  classifier $\bv$ \cite{ilyas2019adversarial}:
\vspace{-5px}
\begin{align}
\nabla_{\widehat{\bh}_t} &=\frac{\partial \ell(\by_t,f(\widehat{\bh}_t;\mathbf{\bv}))}{\partial \widehat{\bh}_t},\\
 \delta &=-\frac{\nabla_{\widehat{\bh}_t}}{\Vert \nabla_{\widehat{\bh}_t} \Vert}_2.\label{pertub}
\end{align}
The gradient $\nabla_{\widehat{\bh}_t}$ carries the information of how to predict $\widehat{\bh}_t$ into class $\by_t$. Therefore, iteratively adding its negative normalized value $\delta$ to $\widehat{\bh}_t$ is towards the direction of loss decreasing of the next optimization iteration.

After obtaining the features $\widehat{\bh}_t$, we cannot guarantee that $\widehat{\bh}_t$ is certainly classified into the target minority class $\by_t$. Therefore, we check out its prediction confidence of $\by_t$ and consider if it can be used for personalized training. We accept all transformed features whose prediction confidence higher than a threshold $p_d$ called \textit{drop probability} in FedAFA. We also empirically validate the influence of $p_d$ in Section \ref{hyper}. After selection by the drop probability, we put the selected generated features $(\widehat{\bh}_t,\by_t)$, as well as the source majority class features $(\bh_s,\by_s)$ into a set $\mathcal{G}_k$ for personalized training.
\vspace{-20px}
	\subsubsection{Personalized Model Optimization} \label{local}  
	One potential risk of using the above feature augmentation method is that it may damage the performance of the source majority classes. While the high-quality generated features of the target minority class $\by_t$ enhance the performance of the local minority classes, some original samples in the source majority class $\by_s$, which are used for augmentation, are likely to be misclassified to the target minority class $\by_t$. This is because the original and generated features $\bh_s$ and $\widehat{\bh}_t$ are still close in the feature space, although they can be classified into different classes. Therefore, to make the classification boundary separate the original and generated features, we proposed a new objective of the local personalized training $\LL_{AFA}$ consisting of two parts: $\LL_{gen}$ and $\LL_{ori}$. They are the averaged training loss on the generated balanced feature set $\mathcal{G}_k^{bal}$ and the class-balanced local dataset $\mathcal{D}_k^{bal}$, respectively:
	\vspace{-10px}
\begin{align}
 \LL_{gen}&=\frac{1}{|\mathcal{G}^{bal}_k|}\sum_{(\bh,\by)\in\mathcal{G}^{bal}_k}\ell(\by,f(\bh;\bv_k)),\label{gene}\\
 \LL_{ori}&=\frac{1}{|\DD^{bal}_k|}\sum_{(\bx,\by)\in \DD^{bal}_k}\ell(\by,f(g(\bx;\mathbf{u});\mathbf{\bv}_k)),\label{origi}\\
 \LL_{AFA}&=\lambda\LL_{gen}+(1-\lambda) \LL_{ori},
\label{local}
\end{align}
where $\lambda$ is a hyperparameter called \textit{balance factor} to control the balance of adjusting the decision boundary between the original and generated features.

\section{Experiments}
\vspace{-5px}

\label{sec:typestyle}
\subsection{Experiment Setup}

\begin{figure}[t]
	\begin{minipage}[b]{0.23\textwidth}
	  \centering
	  \includegraphics[width=\textwidth]{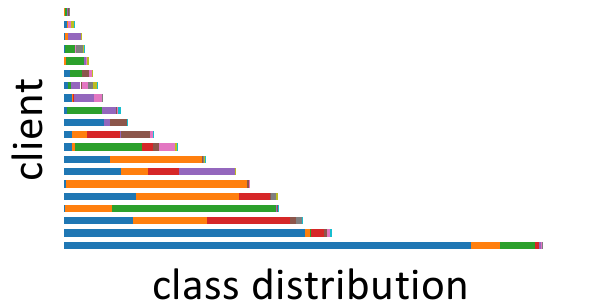}
	
	  \centerline{ (a) CIFAR-10-LT}
	\end{minipage}
	\begin{minipage}[b]{0.23\textwidth}
	  \centering
	  \includegraphics[width=\textwidth]{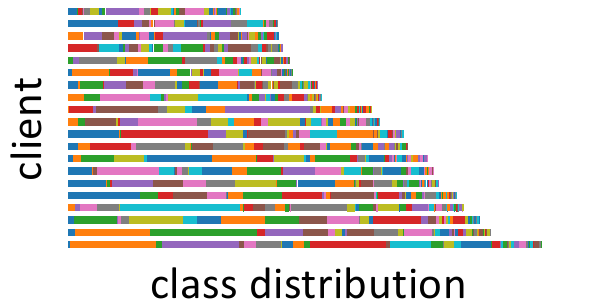}
	  \centerline{(b) CIFAR-100-LT}
	\end{minipage}
	\caption{Class distributions of each client on CIFAR-10/100-LT  with $\alpha=0.2$. Different colors represent different classes, and the length of the color block represents the number of samples in this class.} 
	\vspace{-10px}
	\label{data_dis} 
	\end{figure}

We evaluate FedAFA on two image classification datasets:  CIFAR-10-LT and CIFAR-100-LT \cite{krizhevsky2009learning} \footnote {We refer to the long-tailed versions of these three benchmarks as  CIFAR-10-LT and CIFAR-100-LT.}. Specifically, the number of samples in class $k$ decay exponentially by $\rho^k n_c$ \cite{cui2019class}, where $\rho^k \in (0,1)$ controls the degree of long-tail and $n_c$ is the number of samples in each class of the original balanced dataset. In all experiments, we set the degree of data imbalance at 100, calculated by the number of samples in the largest class divided by the ones in the smallest class. We adopt the Dirichlet Distribution with the hyperparameter $\alpha$ to simulate different degrees of data heterogeneity \cite{hsu2019measuring}. A larger value of $\alpha$ means higher similarity between data distributions. We choose $\alpha=[0.5, 0.2]$ to make heterogeneous data across clients. Figure \ref{data_dis} shows the data distributions of each client for two long-tailed datasets used in our experiments with $\alpha = 0.2$.
We choose ResNet32 as the learning model. The features of the second block of ResNet32 are selected to do feature augmentation of FedAFA. SGD is the optimizer with a learning rate of 0.005, a momentum of 0.9, and a weight decay of $ 5 \times 10^{-4}$ for local model optimization of all methods. We set the total number of clients at 20, the number of clients selected in each round at 10, the batch size at 64, the local epoch for personalized training at 1, and the global communication round at 500.

	\vspace{-10px}

\subsection{Experimental Results}

In this subsection, we first compare FedAFA with two groups of state-of-the-art PFL methods. The methods in the first group are all PFL methods but do not specifically design for imbalanced data. The methods in the other group apply the long-tail learning methods to the PFL framework, including random oversampling (ROS), M2m \cite{kim2020m2m}, and cRT \cite{Kang2020Decoupling}. To verify that the features obtained by the global model are more robust, we also compare FedAFA and FedAFA\_Loc, whose balanced feature sets are generated by the global model and the local personalization model, respectively.
		
\begin{table}[!t]
	
	\small
	\renewcommand\arraystretch{1.2}

	\centering
	\begin{tabular}{l|ll|ll} \hline
		Dataset & \multicolumn{2}{c|}{CIFAR-10-LT} & \multicolumn{2}{c}{CIFAR-100-LT} \\ \hline
  \centering 
  
  $ \alpha$ &0.5&0.2&0.5&0.2\\ \hline
   Local training   & 28.06 & 27.88 & 9.82 & 8.37 \\
  FedAvg-FT \cite{wang2019federated}  & 48.21 &48.11  & 30.66 & 26.32 \\
  FedProx \cite{li2018federated}  & 51.07 &50.26  & 31.40 & 30.58 \\
  LG-FedAvg \cite{l2sgd}  & 51.13 &50.93  & 32.65 & 31.18 \\
  Per-FedAvg \cite{maml}   & 49.96 &49.81  & 32.27 & 30.60 \\
  pFedMe \cite{pfedme}  & 50.05 &49.59  & 32.82 & 30.05\\
  Ditto \cite{li2021ditto}  & 50.84 &50.26  & 31.98 & 31.11 \\
  FedBN \cite{li2021fedbn}  & 50.98 &50.17 & 31.62 & 30.37 \\\hline
  
  FedAvg+ROS   &59.29& 56.51  & 34.04 & 32.02 \\
  FedAvg+M2m  & 62.85 &56.30 & 32.63 & 32.69\\
  FedAvg+cRT  &   {63.33}& 59.55 & {34.42} & {33.11}\\ \hline
  FedAFA\_Loc  &62.24& {62.15}  & 34.12 & 33.06 \\
  
  FedAFA    & \textBF{64.52} & \textBF{63.70}  & \textBF{36.57} & \textBF{35.44} \\ \hline
	\end{tabular}

	\caption{Test accuracy (\%) of FedAFA    compared with different  methods. The best results are shown in bold.}
	\label{table2}
	\vspace{-10px}
	\end{table}

	As shown in Table 1, the performance of local training is poor because the model is only trained locally without any communication with other clients. Other PFL methods that only take data heterogeneity into account without considering the long-tail distribution also underperform compared with the methods that specifically address the long-tail distribution. Although these two groups of PFL methods improve FedAvg-FT from different perspectives, FedAFA is the only method consistently achieving promising personalized performance. In addition, by comparing the test accuracy of FedAFA\_Loc and FedAFA, it is verified that the features extracted by the global model are more robust because more local noise is introduced when selecting the local personalized model to generate features.

\begin{figure}[!t]
	\centering
	\vspace{-5px}
	\includegraphics[width=1\linewidth]{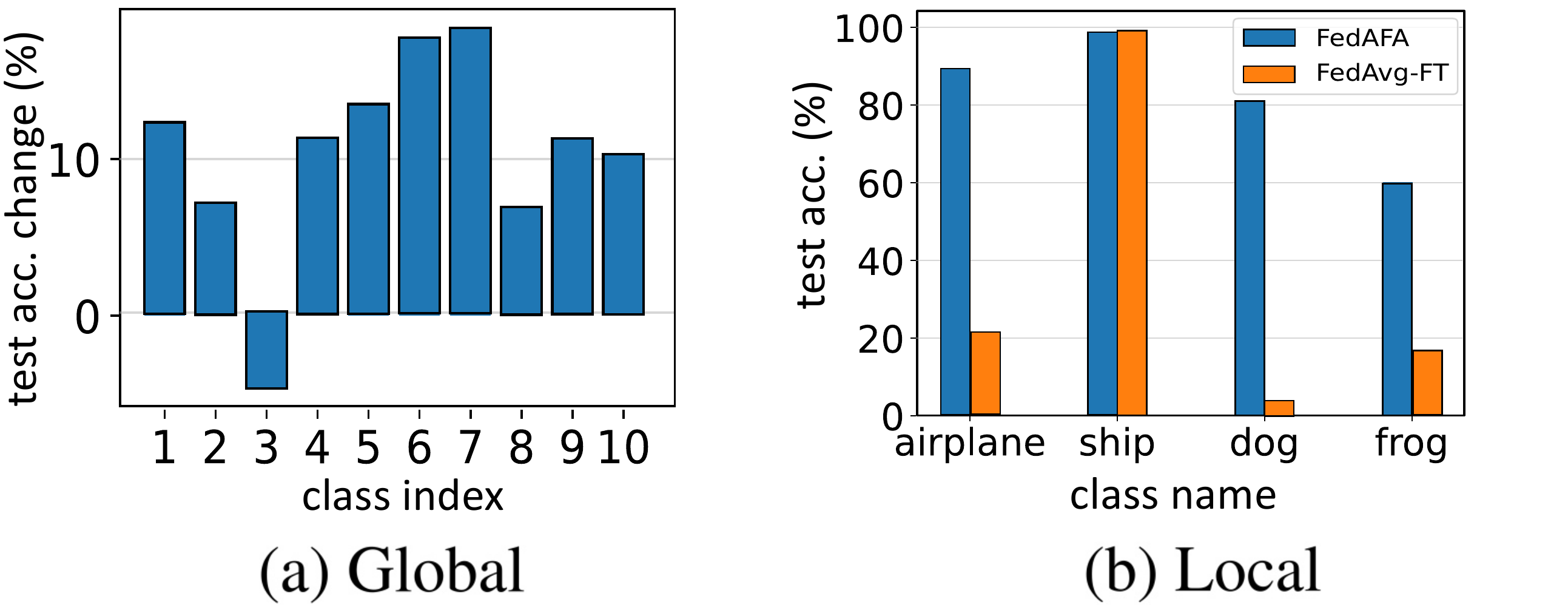}
	\vspace{-20px}
	\caption{Visualization of test accuracy change on each class with $\alpha=0.2$. (a)  The class index is sorted from head classes to tail classes from the global perspective. (b) The numbers of samples of each class from left to right are \{8, 933, 9, 6\}.}\label{acc}
	
	\end{figure}

	To show that FedAFA produces better personalized models for each client, we evaluate the performance boost by FedAFA on each class against the baseline FedAvg-FT from the global perspective in Figure \ref{acc}(a). It can be observed that FedAFA improves the performance for almost every class. In addition, we randomly select one client from all clients to illustrate the improvement of FedAFA on local minority classes. As shown in Figure \ref{acc}(b),  FedAFA can significantly improve the accuracy of local minority classes with only a few samples while keeping the same accuracy on the majority classes. These two experiments validate that FedAFA can improve the tail classes' generalization ability while guaranteeing the head classes' performance.

	\vspace{-7px}
 
\subsection{Effects of Hyperparameters}\label{hyper}
	In FedAFA, there are three hyperparameters: the balance factor $\lambda$, the drop probability $p_d$, and the selected layer for feature augmentation. We conduct various experiments on CIFAR-10-LT to evaluate their influences. 

\begin{figure}
\begin{minipage}[b]{0.15\textwidth}
  \centering
  \includegraphics[width=\textwidth]{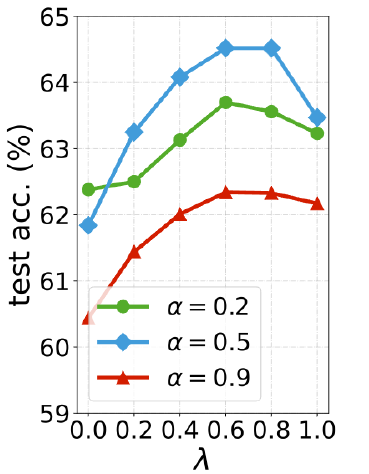}

  \centerline{ (a) $\lambda$}
\end{minipage}
\begin{minipage}[b]{0.15\textwidth}
  \centering
  \includegraphics[width=\textwidth]{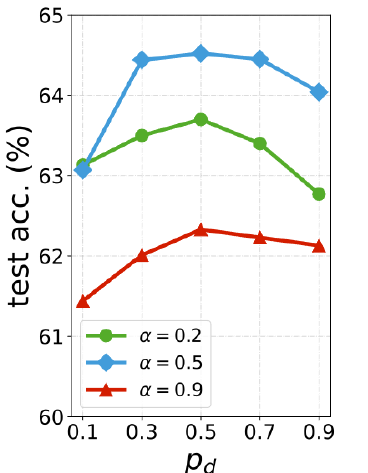}
  \centerline{(b) $p_d$}
\end{minipage}
\begin{minipage}[b]{0.15\textwidth}
  \centering
  \includegraphics[width=\textwidth]{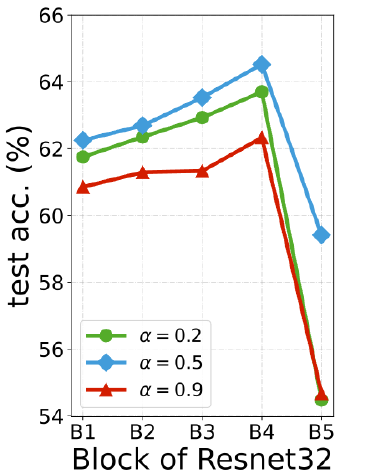}
  \centerline{(c) Selected layer}
\end{minipage}
\vspace{-5px}

\caption{Effects of hyperparameters of FedAFA.} 
\label{hyperparameter} 
\vspace{-15px}
\end{figure}

	We first compare the performance of FedAFA with $\lambda\in [0,1]$  in Figure \ref{hyperparameter}(a). FedAFA degenerates into the case of only training on $\mathcal{D}_k^{bal}$  when $\lambda=0$, and FedAFA is equivalent to the case of only training on $\mathcal{G}_k$ when $\lambda=1$. It can be observed that $\lambda\in[0.6,0.8]$ generally performs better than either extreme case, which validates the effectiveness of the proposed loss function $L_{AFA}$.

	The performance of FedAFA with  $ p_d \in (0,1) $ is also shown in Figure \ref{hyperparameter}(b). It can be observed that the performance increases as $p_d$ reaches 0.5 and then decreases when $p_d$ reaches 1. When $p_d$ is close to 1, only a few generated features can be selected for personalized training, which may only provide limited help. 
	On the other hand, when $p_d$ is close to 0,    generated features with any confidence are used for personalized training. They cannot focus on improving the decision boundary between the local majority and minority classes. Therefore, this experiment validates that drop probability can effectively select generated features helpful for personalized training.
	   
	Finally, Figure \ref{hyperparameter}(c) shows the performance of FedAFA by augmenting features on different blocks of ResNet32. It can be seen that FedAFA performs best when we select the features of Block 3 or Block 4. This observation confirms that selecting the feature space for augmentation is more effective than the sample space.

	\vspace{10px}

\vspace{-20px}

\section{Conclusion}
In this paper, FedAFA is proposed to solve the problem of PFL with heterogeneous and long-tailed data. FedAFA first transfers the knowledge of the local majority class to the local minority class to improve the performance of the local minority class. At the same time, FedAFA proposes a new optimization objective to maintain the performance of local majority classes. Therefore, FedAFA can enhance the generalization ability of the local minority classes while preserving the robust performance of the local majority classes. Experimental results show the superiority of FedAFA compared to other state-of-the-art PFL methods under different settings.

\noindent\textbf{Acknowledgement\ }This study was supported  by the National Natural Science Foundation of China under Grants 62002302 and U21A20514, the Open Research Projects of Zhejiang Lab under Grant 2021KB0AB03, the FuXiaQuan National Independent Innovation Demonstration Zone Collaborative Innovation Platform under Grant 3502ZCQXT2022008 and  the Natural Science Foundation of Fujian Province under Grant 2020J01005.



\vfill\pagebreak

\bibliographystyle{IEEEbib}
\bibliography{strings,refs}

\end{document}